# Hyperspectral and LiDAR data for the prediction via machine learning of tree species, volume and biomass: a contribution for updating forest management plans

Daniele Michelini[1][0000-0002-7995-7985], Michele Dalponte[2][0000-0001-9850-8985], Angelo Carriero[3][0000-0001-8592-5747], Erico Kutchartt[4][0000-0002-9134-4591], Salvatore Eugenio Pappalardo[1][0000-0002-1546-644X], Massimo De Marchi[1][0000-0001-8184-013X], Francesco Pirotti[4,5][0000-0002-4796-6406]

[1] Master in GIScience and UAV, Department of Civil, Environmental and Architectural Engineering (ICEA), University of Padova, Via 8 Febbraio, 2 - 35122 Padova
[2] Research and Innovation Centre, Fondazione Edmund Mach, via E. Mach 1, 38098 San Michele all'Adige, Italy
[3] Servizio Foreste - Provincia Autonoma di Trento, Via G.B. Trener, 3 - 38121 Trento (TN)
[4] TESAF Department, University of Padova, Viale dell'Università, 16 - 35020 Legnaro (PD)
[5] CIRGEO Interdepartmental Research Center in Geomatics, University of Padova, Viale dell'Università, 16 - 35020 Legnaro (PD)

**Abstract.** This work intends to lay the foundations for identifying the prevailing forest types and the delineation of forest units within private forest inventories in the Autonomous Province of Trento (PAT), using currently available remote sensing solutions. In particular, data from LiDAR and hyperspectral surveys of 2014 made available by PAT were acquired and processed. Such studies are very important in the context of forest management scenarios. The method includes defining tree species ground-truth by outlining single tree crowns with polygons and labeling them. Successively two supervised machine learning classifiers, K-Nearest Neighborhood and Support Vector Machine (SVM) were used. The results show that, by setting specific hyperparameters, the SVM methodology gave the best results in classification of tree species. Biomass was estimated using canopy parameters and the Jucker equation for the above ground biomass (AGB) and that of Scrinzi for the tariff volume. Predicted values were compared with 11 field plots of fixed radius where volume and biomass were field-estimated in 2017. Results show significant coefficients of correlation: 0.94 for stem volume and 0.90 for total aboveground tree biomass.

## 1   Introduction

The Autonomous Province of Trento (PAT, the acronym in Italian) has always had a strong forestry vocation, also given the large forest area that it manages: about 390,000 hectares, or 63% of its entire territory. Most of this area is publicly owned:

together with the grazing areas they are part of the forest planning effort (about 397,000 hectares) and are managed in a timely and precise manner through the drafting of local area management plan (LAMP). A small portion of this area (about 77,000 hectares) is grouped into Forest Inventories, which are containers that aggregate small and fragmented private properties that have not had a LAMP since 1995. After 25 years it is necessary to understand what changes have taken place during this time lapse regarding tree species composition, stem volume and aboveground tree biomass distribution, which are information that are requested by PAT legislation [1].

To date, remote sensing technologies are increasingly used in the forestry sector, especially as a support during the back-office phase of field-based surveys. In the case of LAMP, these surveys are mainly used for determining the tree species compositions of the property and thus divide forest areas in forest units by identifying tree species composition and the prevailing height of the stands.

The operations of identifying and classifying the tree species compositions and estimating stem volume and aboveground tree biomass usually take place during the surveys in the field. The collection of information and the drafting of a LAMP therefore involves a substantial financial commitment and working hours in the field. In this work we test the hypothesis that remote sensing, in particular hyperspectral and LiDAR data, can reduce the effort in the field by predicting tree species composition and prevalent tree heights over the study area.

Among the remote sensing technologies mentioned, LiDAR is certainly the most used for predicting stem volume and aboveground tree biomass since the first experiments dating back to 1970. Nowadays LiDAR technology can represent a tool for making accurate measurements, allowing rapid data acquisition to the point of being integrated into the forest planning strategies of many countries[2-3]. The classification of forest species is another aspect that highly relies on hyperspectral images that can be surveyed from airborne platforms, but also more recently unmanned airborne vehicles (UAVs). Hyperspectral data cubes have seen an increase in attention, driven by several scientific studies [4-5, 6].

In various scenarios, hyperspectral sensors can provide data that have shown the possibility of discriminating many tree species with very high precision. Tropical habitats [4], and Mediterranean habitats [7] have shown to be able to distinguish up to 23 different tree species. Also noteworthy is the possibility of being able to discriminate different species of conifers using hyperspectral data [8-9]; in these two investigations, six species of conifers were separated with 70-90% accuracy.

The aim of this specific work is (i) to test the use of hyperspectral data for the identification of the main tree species of Trentino; (ii) integrate LiDAR data into an automatic classification system in order to discriminate forest types and units; (iii) assess volume and biomass estimation from processing LiDAR data.

## 2  Materials and methods

### 2.1  Study area

The areas covered by this investigation are located in the southern part of PAT in Italy and consist of some forest parcels owned by private individuals, located in the municipalities of Calliano, Folgaria, Rovereto and Volano. The privately owned forest stands that do not reach an adequate size to be included in a LAMP are grouped in the inventories of the cadastral municipality in which they are located. These inventories are "management reservoirs" where stand and population information are summarized and, at the same time, where forest cuts made by the owners are communicated, by law, to the PAT Forest Service. The private inventories subject to this analysis are the following: Calliano III and Castelpietra in the Municipality of Calliano, Noriglio in the Municipality of Rovereto, Volano and Folgaria in the Municipalities of the same name (Fig. 1).

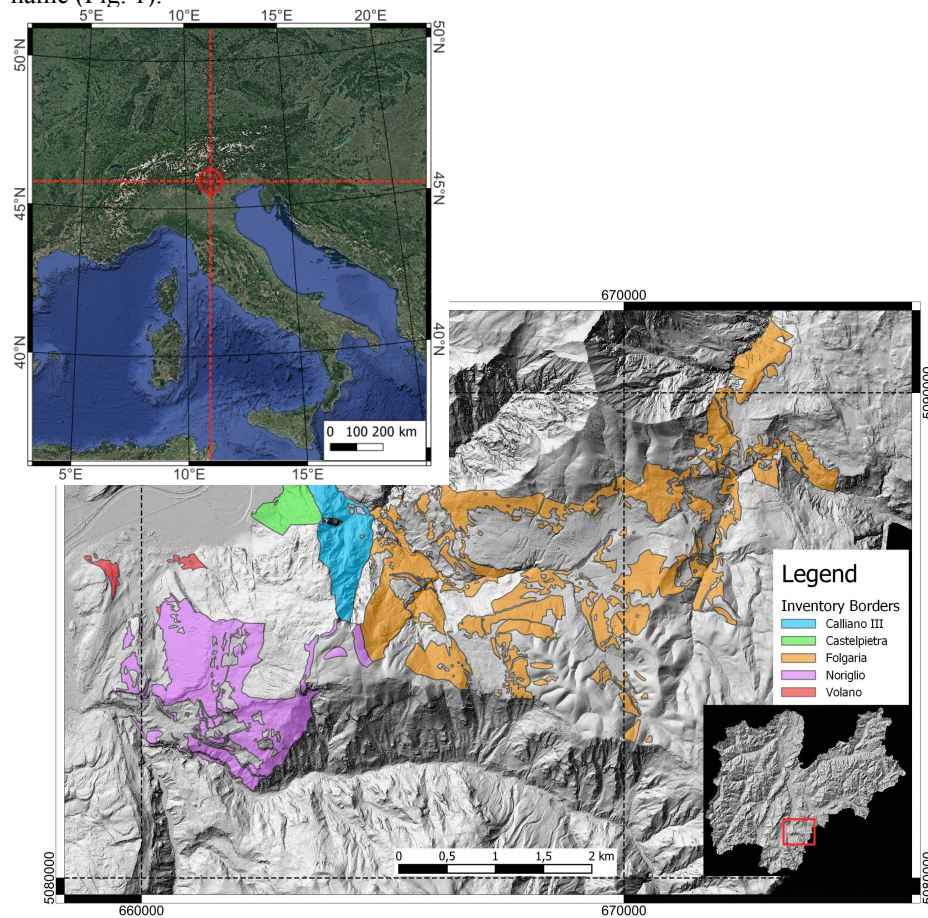

**Fig. 1.** Study area; in the top panel is the overall view with geographic coordinates and the bottom is a detail with coordinate reference system UTM-WGS84 zone 32.

The inventories of Castelpietra, Calliano III and Volano are located entirely in the drainage basin of the Rio Cavallo, while Folgaria is located between the latter, the Centa stream and the Lavarone interzone which, in part, overlooks the Astico stream; finally Noriglio is located between the Rio Cavallo and the Leno stream.

## 2.2 Surveys

The ground surveys consisted of identifying areas to use to train and test the model. These test areas are georeferenced points representing individual trees labeled with species name. Initially 253 trees were defined and geolocated by the PAT Forest Service. Subsequently they were integrated with more surveys in the field, expanding the number up to a total of 891. About 580 individuals (~65%) were used for training the model for species classification and biomass/model estimation and the rest ~35% for testing the model performance using accuracy metrics. Table 1 reports for each species the number for training and test samples.

**Table 1.** List of species and number for samples used for the training and test phases.

| Species name | Training | Test |
|---|---|---|
| Norway spruce (*Picea abies* Karst.) | 115 | 61 |
| Silver fir (*Abies alba* Mill.) | 53 | 24 |
| Larch (*Larix decidua* Mill.) | 78 | 43 |
| Scots pine (*Pinus sylvestris* L.) | 60 | 30 |
| Black pine (*Pinus nigra* Arn.) | 31 | 15 |
| Beech (*Fagus sylvatica* L.) | 71 | 39 |
| Downy oak (*Quercus pubescens* Willd.) | 17 | 8 |
| Hop-hornbeam (*Ostrya carpinifolia* Scop.) | 28 | 18 |
| Manna ash (*Fraxinus ornus* L.) | 14 | 9 |
| European ash (*Fraxinus excelsior* L.) | 12 | 7 |
| Sycamore (*Acer pseudoplatanus* L.) | 16 | 12 |
| Birch (*Betula pendula* Roth) | 10 | 10 |
| Turkey oak (*Quercus cerris* L.) | 3 | 2 |
| Other conifers | 12 | 5 |
| Other broadleaves | 60 | 28 |

The survey includes fifteen tree species which are not homogeneously distributed within the area. Some species are under-represented and some over-represented due both to the specialization of some species in certain ecological niches (*e.g.*, *Abies alba*), and to the difficulty of finding suitable individuals for the ground truth (intertwined foliage and rarity of the species). To compare the predicted biomass and volume, data deriving from private inventories made in 2017 by the PAT were used.

**2.3 Remote sensing data**

The PAT LiDAR data were acquired with the ALT M Gemini laser-scanning system (ALTM 3100 EA and LMS Q780) with an average point density of 10 pts/m$^2$. From this survey three products are available and were used in this investigation: the raw point cloud, the Digital Terrain Model (DTM) and the Canopy Height Model (CHM).

The hyperspectral survey was carried out with a CASI 1500 sensor from the Itres manufacturing company, capable of acquiring bands in the visible and near infrared (VNIR) range of the electromagnetic spectrum, from 0.38 to 1.05 μm with up to 288 spectral channels or bands, with a Field of View (FOV) of 40°. It is a pushbroom sensor with an across-track size of 1500 pixels. The survey campaign was carried out in fifteen days of flight that resulted in 268 strips that were processed to 39 batches of contiguous areas.

The mosaic process was carried out with the following steps: (i) radiometric calibration, (ii) mosaicking and normalization. Mosaicking the tiles was carried out using the R programming environment. Due to the excessive spectral noise, the first 7 bands and the last 8 bands were eliminated. The radiometric calibration was considered using a single strip as reference and all bands were analyzed, band by band. The mosaicking was carried out using the Mosaicking function of the ENVI® software and, in particular, without carrying out any further calibration. The last step was to normalize the data to reduce the differences in mosaic illumination and at the different times the strips were acquired. The normalization affected each pixel of the image based on the average of the spectral signature of the pixel itself:

$$x_{new} = \frac{x}{\frac{1}{N}\sum_{i=1}^{N} x_i} \tag{1}$$

where $x$ is the value of the pixel to normalize, $x_{new}$ is the value of the normalized pixel, $x_i$ is the value of the pixel to normalize on the band $i$, and $N$ is the number of image bands. ENVI® software was also used for normalization and in particular the sum data bands and band math operations.

**2.4 Soil morphology analysis**

The analysis of the local topography (elevation, slope and aspect) was essential to determine the prevailing forest type, for example the Altimontana xerica spruce forest. The height above sea level was calculated by analyzing the DTM and dividing it into altitude classes using the QGIS Raster calculator, while the slope and exposure were determined starting from the raster analysis tools of QGIS.

**2.5 Delineation of tree crowns**

The CHM is a raster image that represents the heights of the foliage. These images are extrapolated from LiDAR data or a normalized digital terrain surface model (nDSM) and are typically used in the forestry field to collect information useful for forest management [10-11]. The CHM is also used as a basis for many important calculations, so its accuracy is of fundamental importance. An example of these applications is represented by the biomass estimation [12] and the specific composition [13], some authors have obtained good results starting from the extraction of the "model trees" from a CHM. Others, on the other hand, started from CHM for the delineation of stands [14] and for the identification of suitable habitats for fauna species [15].

In this work, starting from LiDAR data in LAS format, the CHM was extracted using the *lidR* library in R program. The function used is *grid_canopy* which is based on the *pitfree* algorithm. The process involves the creation of several triangulations at different heights from which different CHMs are extracted which are then combined to generate the final CHM [2]. The most recent LiDAR surveys have a density of points such as to allow the extraction of individual tree crowns (ITC) and there are numerous methods to do this [12, 16-19].

In this case study we opted for the *itcLiDAR* approach [12] inside the *itcSegment* library within the R environment. This library is able to extract the ITCs starting from a LAS file. This approach identifies the treetops within a raster CHM and from there defines the individual crowns around the highest point of the canopy. The same parameters were used throughout the survey area (Table 2).

**Table 2.** Parameters of the *itcLiDAR* algorithm used in this study.

| Variable | Value | Variable | Value |
|---|---|---|---|
| resolution | 0.5 | TRESHSeed | 0.55 |
| MinSearchFilSize | 3 | minDIST | 5 |
| MaxSearchFilSize | 7 | maxDIST | 40 |
| TRESHCrown | 0.6 | HeightThreshold | 2 |

### 2.6 Spectral feature selection

Numerous studies have reported that, with regard to the characterization of a forest stand, it is more important the position of the bands in the spectrum, with respect to the number of spectral bands [20,21]. Some authors have reported how the classification of tree species is facilitated by using bands in the near infrared, subject to a lower amount of noise due to the atmospheric effects of the medium infrared [4,22] and in the red bands [20,23].

The hyperspectral survey used in this investigation is characterized by 122 bands distributed over all the visible and near infrared spectrum and thus for this reason, before moving on to the classification phase, a feature selection operation was performed. In this study, a feature selection method based on the Sequential Forward Floating Selection (SFFS) algorithm and the Jeffries-Matusita distance [24,25] was used to identify the set of the suboptimal bands. Once the spectral bands to be used were identified, the actual classification was carried out.

## 2.7 Classifiers

In this study we used two non-parametric supervised machine learning classifiers: (1) K-Nearest Neighborhood and (2) Support Vector Machine (SVM).

The K-Nearest Neighborhood supervised non-parametric classification algorithm defines, starting from the training data, the average values of each class in the n-dimensional space defined by the input features. The result is an average of the classes with as many values as there are classes identified in the training data. The Euclidean distance of each pixel from the various centroids of the classes defined by the training areas is then calculated; finally, the pixels are attributed to the class whose centroid is placed at the minimum distance [26]. The SVM algorithm (supervised non-parametric classification algorithm) is based on the principle that the space of the starting features can be transformed into a higher-dimensional space, in which the classes are linearly separable. The transformation is performed using a Gaussian-type kernel function, the Radial Basis Function. As for the previous algorithm, the hyperspectral raster file, the ground truth vector file and the list of bands to be used were used. The raster has also been classified several times with different cost parameters $C$.

The starting data for the classification are the hyperspectral images, that are in raster stack format, and a polygonal vector file of the ground truths where each polygon geometry is labeled with the species and are used as training data and the list of bands to be used for the classification. The vector file of the truths on the ground was obtained starting from the file of the canopy polygons extracted by canopy delineation done in the previous step. The file was validated using ground surveys that geo-positioned trees with a Global Navigation Satellite System (GNSS) and assigned tree species on the ground. The point file was then brought into the QGIS environment, and a spatial join with the delineated ITCs polygons provides the species' information to the training and test ITCs.

## 2.8 Biomass and volume estimation models

The biomass estimate was calculated starting from the LiDAR survey and in particular from the normalized LAS point cloud with which the crowns were extracted using the itcLiDAR R package; together with the extension of the tree canopy crowns, the package also reports the area and tree height for each of them. For the actual calculation, two equations were used that identify aboveground biomass (AGB) and volume. These were compared using the allometric Jucker equation [27] and the Scrinzi tariff volume equation [28]. Both equations are implemented inside the R libraries mentioned above. These two equations, as well as all existing allometric calculations, are based on the tree diameter as the basis for the volume estimate. Since this cannot be measured directly from the LiDAR survey, it was estimated using the equation identified by Jucker [27]. This equation takes into account, in addition to the height, also the diameter of the crown. In fact, although there is strong correlation between height and crown diameter, the relationship between these two variables varies greatly both within the species and as a function of climate and structure [29-30-31, 32] .

It should also be noted that, while taking these parameters into consideration, it is complex to identify a unique equation [33-29-34-35, 36]. In most cases, trees grow very tall in order to capture as much light as possible until, once they reach the maximum height, they develop above all the diameter. Hence, a group of large individuals can have very different diameters. The relationship between crown size and diameter, on the other hand, tends to remain more constant over time as the stem of the plant must continue to grow to maintain stability and water supply [37-38, 35], even after reaching the maximum height. For this reason, these two parameters are strongly correlated even in trees with large diameters [39].

AGB is a parameter strongly correlated to height and crown diameter (H × CD) which sees a great variation between functional groups, that is between gymnosperms and angiosperms. The most suitable equation identified, once taking into account the different relationships between angiosperms and gymnosperms, is the following:

$$AGB_{predicted} = (0.016 + \alpha_G) \times (H \times CD)^{(2.013+\beta_G)} \times exp\left(\frac{0.204^2}{2}\right) \qquad (2)$$

where $\alpha_G$ and $\beta_G$ are functional group-dependent parameters representing the difference in scale constant $\alpha$ and scale exponent $\beta$ between angiosperms and gymnosperms (gymnosperms: $\alpha_G = 0.093$ and $\beta_G = -0.223$; angiosperms: $\alpha_G = 0$ and $\beta_G = 0$). The Scrinzi tariff volume equation, on which the second estimation method is based, is as follows:

$$V = b_0 + b_1 G + b_2 GP_s + b_3 GP_s I_t + b_4 GP_s B_d \qquad (3)$$

where $G$ is the basal area per hectare, $b_0, b_1, b_2, b_3$ and $b_4$ are the regression coefficients, $P_s$ is the stereometric potential index of the species, $I_t$ is the tariff index, and $B_d$ is the barycentric dimensional index.

The authors also developed the equation for estimating the volume of the tree trunk to be used in a double entry table:

$$V = a \times (d-d_0)^b \times h^c \qquad (4)$$

where $h$ is the total height of the plant and $d$ the diameter at breast height while $a, b, c$ and $d_0$ are parameters that depend on the different tree species and are reported in the table below:

Table 3. Parameters for equation for the main tree species (equation 4).

| P. | Tree species | | | | | | |
|---|---|---|---|---|---|---|---|
|  | *P. abies* | *A. alba* | *L. decidua* | *F. sylvatica* | *P. silvestris* | *P. cembra* | *P. nigra* |
| a | 0.000177 | 0.000163 | 0.000108 | 0.000055 | 0.000102 | 0.000188 | 0.000129 |
| b | 1.564254 | 1.706560 | 1.407756 | 1.942089 | 1.918184 | 1.613713 | 1.763086 |
| c | 1.051565 | 0.941905 | 1.341377 | 1.006420 | 0.830164 | 0.985266 | 0.938445 |
| $d_0$ | 3.694650 | 3.694650 | 3.694650 | 4.009100 | 3.694650 | 3.694650 | 3.694650 |

## 3 Results

### 3.1 Determination of the species

From the feature selection process 35 bands of interest were identified for classification. Figure 2 shows an extract of the cartography resulting from three of the different classifications used:

   a) K-Nearest Neighborhood
   b) Support Vector Machine with all surveyed species;
   c) Support Vector Machine with aggregations of species: we analyzed only the species that are subject to forest planning management (spruce, silver fir, larch, scots pine, black pine and beech) while the rest were grouped into three forest types (ornoo-ostrio-oak, maple-ash, other broad-leaved trees and other conifers).

The first visible evidence in Figure 2 (panel A) is the presence of a gap that cuts the map obliquely; this is an area where in the hyperspectral images some bands had problems. These bands are variable and usually range from band 1 (0.39640 µm) to band 10 (0.43940 µm). In many analyzes, the bands of that range extracted from the feature selection have been manually eliminated (in this case we went from 35 to 28 bands). By eliminating these bands, the edge effect between one stripe and the next was also limited, as visible in the other two panels. The edge effect is due to the time elapsed between two stripes acquisitions, that sometimes is very long (a morning).

Panel C (Figure 2) shows the result obtained by the SVM classifier with the parameter $C$ set to 10, by carrying out a test based on the classification of only the most important species in Trentino and grouping the others into similar species. Table 4 shows the main accuracy metrics for the classifiers used. Accuracy is defined as the ratio of correct values (true positives) over the total number of values, and precision is defined as the ratio of correct values (true positives) over all values for a specific class, including false . The F-score measures the accuracy of the test that takes into account precision and recall (the number of true positives divided by the number of true positives plus false negatives).

The best results are those related to the classification carried out with SVM and its parameter $C$ set to 10, both for the tree species being analyzed and for the other metrics identified. The classification accuracy of the most important species from the point of view of forest management range from 96% for black pine to 73% for beech; even firs (spruce: 85%, silver fir: 90%) performed well. Good precision (91%) was also obtained for the black pine. Lower performances resulted in the identification precision values of silver fir (55%), often confused with red fir. The F-score, on the other hand, goes from 48% for beech to 71% for black pine while most of the other species still reach a satisfactory score: spruce 70%, silver fir 62% and larch 65%.

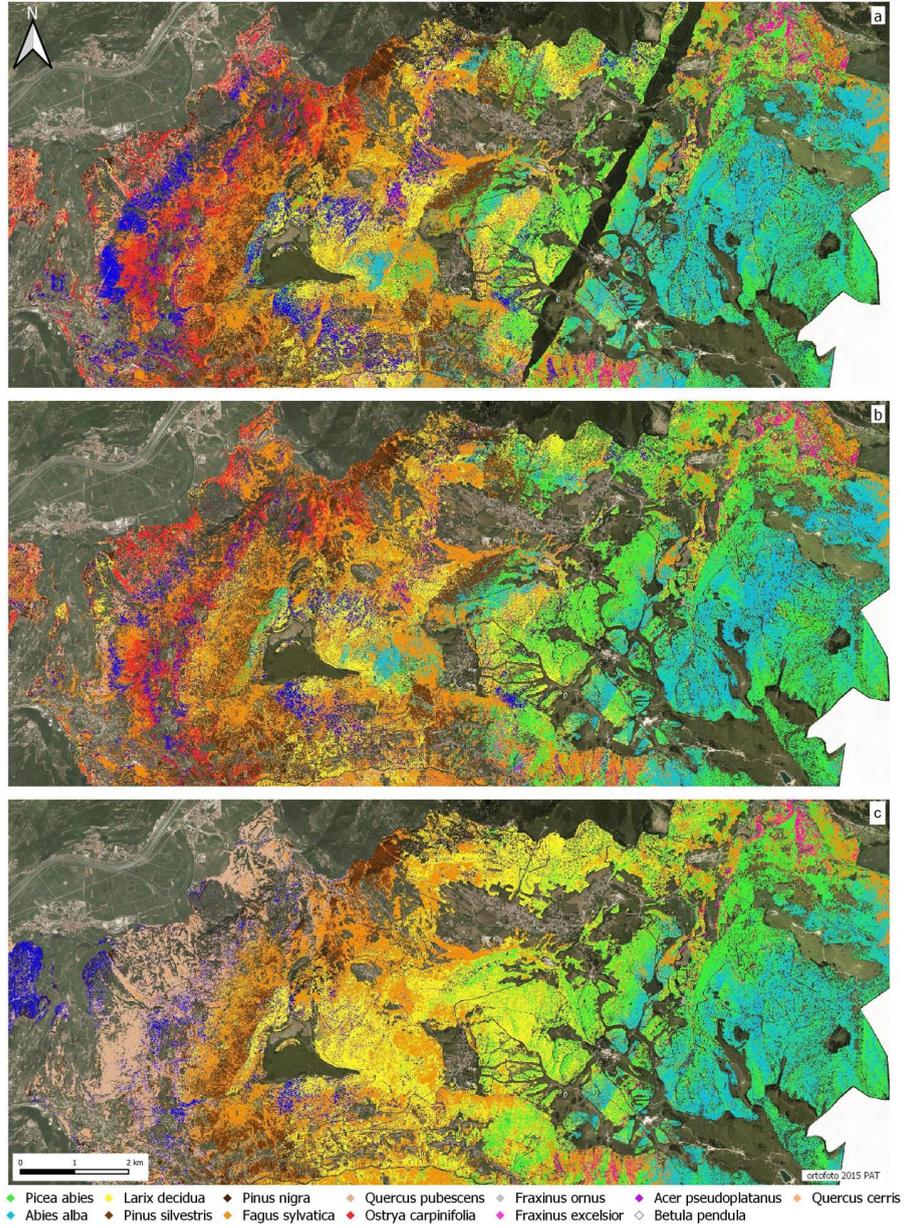

**Fig. 2.** Hyperspectral classification comparison: (a) K-Nearest Neighborhood; (b) SVM with all surveyed species; (c) SVM with aggregations of species.

**Table 4.** Accuracy metrics from test data sets. Acc.=accuracy, Prec.=Precision.

| Species | K-means | | | SVM all species | | | SVM subset of species | | |
|---|---|---|---|---|---|---|---|---|---|
| | Acc. | Prec. | F | Acc. | Prec. | F | Acc. | Prec. | F |
| *P. abies* | 78% | 64% | 56% | 85% | 75% | 70% | 69% | 53% | 52% |
| *A. alba* | 85% | 44% | 55% | 90% | 55% | 62% | 80% | 39% | 41% |
| *L. decidua* | 76% | 44% | 50% | 85% | 59% | 65% | 65% | 33% | 42% |
| *P. silvestris* | 78% | 37% | 42% | 82% | 42% | 49% | 73% | 32% | 32% |
| *P. nigra* | 94% | 63% | 69% | 96% | 91% | 71% | 91% | 60% | 60% |
| *F. sylvatica* | 74% | 64% | 49% | 73% | 51% | 48% | 65% | 54% | 47% |
| *Q. pubescens* | 92% | 17% | 12% | 95% | 100% | 44% | 77% | 0% | - |
| *O. carpinifolia* | 85% | 63% | 45% | 88% | 73% | 44% | 70% | 36% | 32% |
| *F. ornus* | 92% | - | - | 93% | 100% | 12% | 76% | 7% | 9% |
| *F. excelsior* | 92% | 0% | - | 93% | 25% | 13% | 89% | 17% | 12% |
| *A. pseudoplatanus* | 90% | 83% | 34% | 92% | 83% | 36% | 88% | 67% | 31% |
| *B. pendula* | 94% | - | - | 94% | 100% | 33% | 82% | 15% | 17% |
| *Q. cerris* | 97% | 40% | 44% | 99% | 100% | 67% | 86% | 11% | 16% |
| Other conifers | 99% | 100% | 67% | 99% | 100% | 75% | 97% | 100% | 33% |
| Other broadleaves | 83% | 68% | 51% | 84% | 70% | 55% | 78% | 61% | 49% |

### 3.2 Estimation of volume and above ground biomass

The above ground biomass and volume equations are species-specific, therefore each ITC that was delineated in the initial steps is labeled with a species using the results from the classification step. To do this the majority statistic within each single ITC is used over all classified pixels inside the ITC. The specific equation was then applied to each tree/canopy to calculate tree-based AGB and stem volume.

At the end of the process tree AGB/volume values, expressed in m$^3$ and a biomass value, expressed in kg, were aggregated for each area. The biomass represents the dendrometric biomass above ground (AGB). In order to verify these data, as reference data we used the fixed-area surveys that were carried out on behalf of the PAT on all private forests in Trentino. These are surveys on circular areas of 15 m radius where all tree individuals with a trunk diameter at breast height (DBH) greater than 7.5 cm have been measured for height and diameter and the equations (2-4) used to calculate volume and AGB. The number of parcels that were used to compare ground-truth is relatively small (11) but provides a rigorous comparison. The eleven areas were also scattered and they are also at a great distance from each other and forest compositions are also quite different. This is also visible from the great heterogeneity of the quantities estimated and reported in Table 6.

**Table 6.** Biomass and volume results from observed ground truth values (Ob) and estimated predicted values (Pr) for volume (V) and above ground biomass (AGB).

| | Area | 1 | 2 | 3 | 4 | 5 | 6 | 7 | 8 | 9 | 10 | 11 |
|---|---|---|---|---|---|---|---|---|---|---|---|---|
| V (m$^3$) | Ob | 3.00 | 3.33 | 57.87 | 21.32 | 15.41 | 12.22 | 14.53 | 25.09 | 18.42 | 6.74 | 20.51 |
| | Pr | 1.02 | 1.32 | 63.92 | 13.14 | 12.22 | 3.19 | 5.67 | 12.40 | 11.57 | 0.34 | 0.23 |
| AGB (Mg) | Ob | 1.84 | 1.99 | 26.95 | 11.82 | 7.49 | 6.10 | 7.24 | 12.76 | 9.21 | 4.04 | 12.25 |
| | Pr | 1.07 | 1.45 | 37.01 | 8.99 | 9.19 | 2.68 | 5.39 | 9.53 | 7.43 | 5.48 | 3.68 |

Finally, in Figure 3, the scatter plots are shown which show a good correlation between the data, even if at times significantly different. This is also proved by the correlation coefficient (R): 0.94 for the stem volume and 0.90 for the aboveground tree biomass.

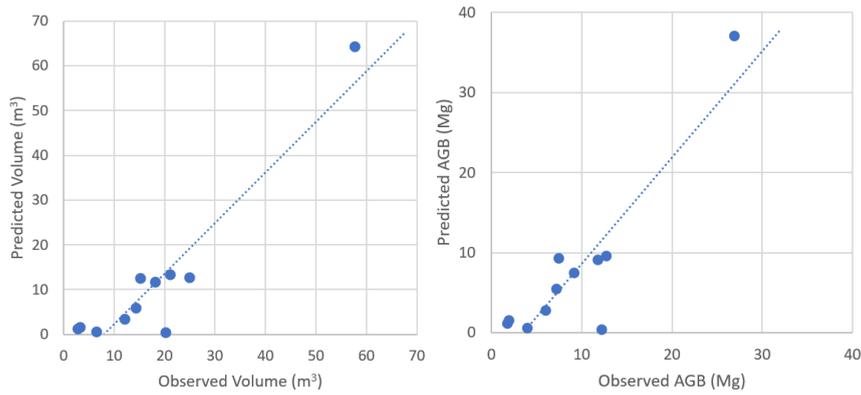

**Figure 3.** Predicted vs. observed stem volume and aboveground biomass scatter plots respectively with R=0.94 and R=0.90.

## 3   Discussion and conclusions

Comparing the classifiers used, SVM is by far the one that produces the most reliable results in this study. Other machine learning methods such as random forest have proven also to perform very well, with some advantages in many cases not only in forest environments [40-42]. Considering the distribution of the main species analyzed, Norway spruce is the one that reaches the best accuracy, notably thanks to the greater number of field samples used to train the model. The silver fir, dominant only in the area where the hyperspectral relief shows some defects, is very susceptible to false positives towards the spruce class, while maintaining a very high accuracy. Larch has a high number of false positives, which affects its accuracy and F-score. The determination of the Scots pine shows excellent results in Noriglio inventory area, while in the Folgaria area it is less efficient also with respect to the larch class. For the black pine, on the other hand, the classifier provides better results despite the fact that the number of ground-truths is not very high. As the beech tree is the most present and easiest to identify broadleaf tree, it appears to have numerous false negatives, especially regarding the analysis in the total area, which affects its accuracy.

The estimate of the tree biomass and the tariff volume is good despite the limited number of comparison areas and the parameters set for the determination of the foliage. A very important aspect to take into consideration is precisely the parameters with which the tree canopy crown of the plants were extracted. Given the extent of the area, we used parameters that would be general and adapt to different scenarios. It is reasonable to hypothesize that this brings higher errors than using site-specific parameters, in the estimated volume and biomass from LiDAR data.

The objective of this study was to outline the main features needed to identify the forest units of some Forest Inventories and to provide an overview of the biomass present starting from some remote sensing data available over PAT. The main conclusions that emerge from this work are:

(i) the presence of a high point density LiDAR survey is fundamental for a good result of the processing [43]: the identification of the canopy height with as little error as possible allows not only the estimation of biomass and volumes but also a correct classification of the species - approaches like full-waveform metrics [44] can provide in some cases improvements, but are more complex to process;

(ii) among the two classifiers used only SVM gave optimal results; the survey carried out on an area that is too large generates excessive noise in the model due to a frequency difference in the incident and reflected electromagnetic waves, the same thing can be said in the case of surveys carried out with orthogonal streaks to the slopes and too distant in time;

(iii) among the species analyzed, the one that resulted easier to identify is certainly the spruce, probably also thanks to the widespread diffusion of this species; the choices of individual trees for the field data is fundamental and they cannot always be found in adequate numbers for all the species (*e.g.*, silver fir and hornbeam) or in optimal conditions (crowns intersected or subjected at the time of the hyperspectral survey).

## Author Contributions

All authors contributed to the study conception and design. Michelini designed and carried out the methodology, and wrote the first draft of the paper. Dalponte and Carriero supported conception and data collection. Kutchartt carried out reviews and extensive editing. Pappalardo and De Marchi supervised the conceptual design and reviewed the paper. Pirotti reviewed the first draft and supervised the methodological approach. All authors commented on previous versions of the manuscript and have read and approved the final manuscript.